\documentclass[sigconf]{acmart}
\AtBeginDocument{%
  \providecommand\BibTeX{{%
    \normalfont B\kern-0.5em{\scshape i\kern-0.25em b}\kern-0.8em\TeX}}}

\copyrightyear{2022}
\acmYear{2022}
\setcopyright{acmlicensed}
\acmConference[KDD '22] {Proceedings of the 28th ACM SIGKDD Conference on Knowledge Discovery and Data Mining}{August 14--18, 2022}{Washington, DC, USA.}
\acmBooktitle{Proceedings of the 28th ACM SIGKDD Conference on Knowledge Discovery and Data Mining (KDD '22), August 14--18, 2022, Washington, DC, USA}
\acmPrice{15.00}
\acmISBN{978-1-4503-9385-0/22/08}
\acmDOI{10.1145/3534678.3539446}

\usepackage[ruled,linesnumbered]{algorithm2e}
\usepackage{multirow}
\usepackage{siunitx}
\usepackage{physics}
\usepackage{tabularx}

\usepackage{etoolbox}
\newrobustcmd\B{\DeclareFontSeriesDefault[rm]{bf}{b}\bfseries}    

\begin{document}

\title{RetroGraph: Retrosynthetic Planning with Graph Search}


\author{Shufang Xie}
\affiliation{
    \institution{Gaoling School of AI (GSAI) \\ Renmin University of China }
    \country{}
}
\email{shufangxie@ruc.edu.cn}

\author{Rui Yan}
\authornote{Corresponding author: Rui Yan (ruiyan@ruc.edu.cn), Peng Han (pengh@cs.aau.dk)}
\affiliation{
    \institution{Gaoling School of AI (GSAI) \\ Renmin University of China }
    \country{}
}
\email{ruiyan@ruc.edu.cn}

\author{Peng Han}
\authornotemark[1]
\affiliation{
    \institution{Aalborg University}
    \country{}
}
\email{pengh@cs.aau.dk}

\author{Yingce Xia}
\affiliation{
    \institution{Microsoft Research Asia}
    \country{}
}
\email{yingce.xia@microsoft.com}

\author{Lijun Wu}
\affiliation{
    \institution{Microsoft Research Asia}
    \country{}
}
\email{lijuwu@microsoft.com}

\author{Chenjuan Guo}
\affiliation{
    \institution{East China Normal University}
    \country{}
}
\email{cjguo@dase.ecnu.edu.cn}

\author{Bin Yang}
\affiliation{
    \institution{East China Normal University}
    \country{}
}
\email{byang@dase.ecnu.edu.cn}
\author{Tao Qin}

\affiliation{
    \institution{Microsoft Research Asia}
    \country{}
}
\email{taoqin@microsoft.com}

\renewcommand{\shortauthors}{Shufang, et al.}


\newcommand\vertset{\mathcal{V}}
\newcommand\edgeset{\mathcal{E}}
\newcommand\rec{\text{rec}}
\newcommand\molt{\text{mol}}
\newcommand\success{\text{succ}}
\newcommand\cost{\text{cost}}
\newcommand{\vnext}{v_{\text{next}}}

\begin{abstract}
Retrosynthetic planning, which aims to find a reaction pathway to synthesize a target molecule, plays an important role in chemistry and drug discovery. This task is usually modeled as a search problem.
Recently, data-driven methods have attracted many research interests and shown promising results for retrosynthetic planning.
We observe that the same intermediate molecules are visited many times in the searching process, and they are usually independently treated in previous tree-based methods (e.g., AND-OR tree search, Monte Carlo tree search). Such redundancies make the search process inefficient. We propose a graph-based search policy that eliminates the redundant explorations of any intermediate molecules. As searching over a graph is more complicated than over a tree, we further adopt a graph neural network to guide the search over graphs.
Meanwhile, our method can search a batch of targets together in the graph and remove the inter-target duplication in the tree-based search methods.
Experimental results on two datasets demonstrate the effectiveness of our method.
Especially on the widely used USPTO benchmark, we improve the search success rate to 99.47\%, advancing previous state-of-the-art performance for 2.6 points.
\end{abstract}




\maketitle

\newcommand{\uspto}{\texttt{USPTO}}
\newcommand{\usptoext}{\texttt{USPTO-EXT}}

\section{Introduction}
\begin{figure}[!htbp]
\centering
\includegraphics[width=\linewidth]{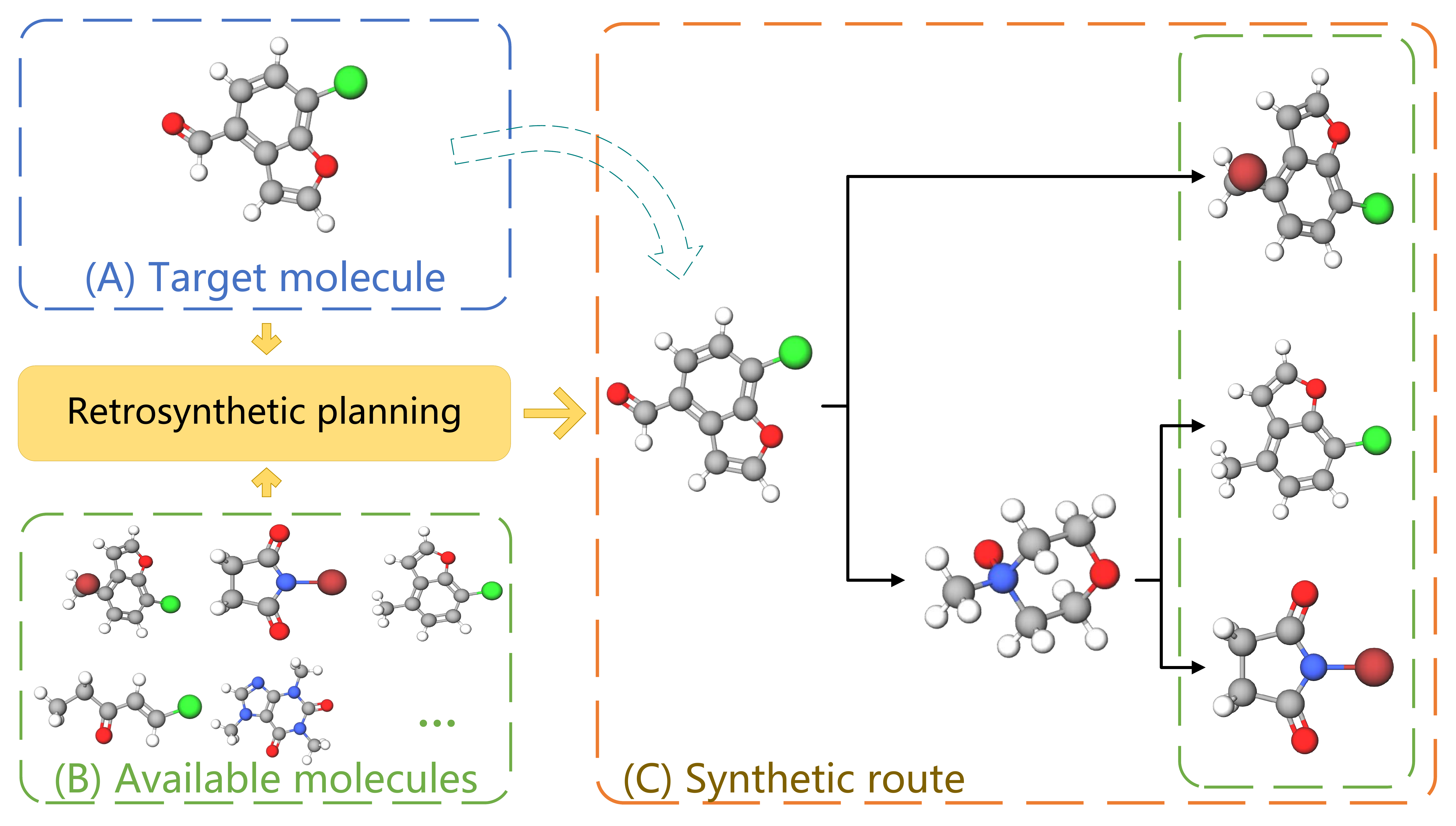}
\caption{Example of retrosynthetic planning. Given a target molecule (A) and a set of available molecules (B), the retrosynthetic planning system is to find a feasible route (C).}
\label{fig:intro}
\end{figure}

Retrosynthetic planning aims to find pathways to synthesize novel molecules and is an important topic in chemistry and drug discovery.
As shown in Figure~\ref{fig:intro}, given a target molecule and a set of ingredient molecules, the goal is to find a pathway where the target can be eventually synthesized using ingredient molecules, and each step of the pathway is a viable chemical reaction. Such a process is usually modeled as a tree-search problem, where the starting point is the target molecule, and the path is the chemical reaction.
Research focus~\cite{Chen2020,Kim2021,Hong2021} is how to find a feasible pathway with as few trials as possible.

Various data-driven search policies have been leveraged / proposed for this task, such as A* search~\citep{Chen2020}, proof number search~\citep{kishimoto2019depth}, and Monte Carlo tree search~\citep{Hong2021}. They treat the planning progress as a tree expansion problem. 
However, as shown in Section~\ref{sec:res_study}, the success of those methods is limited  in retrosynthetic planning due to intra-target and inter-target redundancy in search trees:

(1) The search tree of one target molecule could have multiple identical subtrees, because similar reactions can produce the same intermediate molecules. For example, known as Finkelstein Reaction~\citep{finkelstein1910darstellung}, \textit{iodoethane} can be synthesized by \textit{bromoethane + podium iodide} or \textit{bromoethane + potassium iodide}. Therefore, in the search tree of \textit{iodoethane}, the whole subtree of \textit{bromoethane} will be expanded generated multiple times in previous work~\cite{Chen2020,Kim2021} , making the search inefficient, which is referred as \emph{intra-target} redundancy. 

(2) Previous methods~\cite{Chen2020,Kim2021} treat different targets separately while ignoring that they may share some common intermediate molecules after several reactions, which is common in synthetic chemistry.
For example, \textit{6-aminopenicillanic acid (6-APA)} is a common intermediate for a lot of \textit{$\beta$-lactam antibiotics} such as \textit{benzylpenicillin}, \textit{amoxycillin}, and \textit{methicillin}~\cite{batchelor1959synthesis}.
Previous methods repeatedly visit the same intermediate molecules and their sub-trees while planning for related target molecules, which results in \emph{inter-target} redundancy.

Therefore, we propose to merge duplicated molecule nodes in the search trees to eliminate the intra-target and inter-target redundancy.
More specifically, instead of modeling the search process by a tree, we use a directed graph to represent the search process.
The graph has two kinds of nodes: molecule nodes and reaction nodes. A molecule node represents an unique molecule which can be a target molecule, an intermediate reactant, or an ingredient molecule. The children of a molecule node are possible reactions that can synthesize the molecule in one step. The reaction node represents a reaction, and its children are its corresponding reactants.

A technical challenge of searching over a retrosynthetic planning tree is that the selection of one intermediate molecule is affected by its sibling sub-trees due to the existence of reaction nodes.
In an extreme case, if one reactant cannot be synthesized from a given set of ingredients, we do not want to waste trials on all the reactions involving this reactant.
The problem becomes more challenging when we change the search structure from tree to graph because each molecule can have multiple predecessors (i.e., a molecule is a reactant of multiple reactions).
Although we can design heuristic rules to handle this extreme case, making the rules complete and optimal is difficult.
Instead, we propose a learning-based method that uses a graph neural network (GNN) to guide search. This GNN takes a whole search graph as input and outputs a score for each molecular node in the graph to indicate the likelihood of the success of synthesizing a molecule.
Because the whole graph information is available to the GNN, we expect it can better suggest the next node to expand.

Overall, the searching algorithm contains three steps: (a) selecting the most promising molecule to expand based on the known cost and GNN prediction of future cost; (b) expanding the selected node with a single step retrosynthetic network. Meanwhile, we merge duplicated molecule nodes in the search graph; (c) updating the nodes' properties (e.g., whether a node is success) for the next round. Within the exploration budget, we repeat such a process until we find retrosynthetic plans for all target molecules.

Furthermore, using the graph-search method, we can naturally process a batch of targets molecules together by building a hyper-graph that common ingredients are shared, i.e., there are no duplicated molecule nodes in the hyper-graph. 

To demonstrate the effectiveness of our algorithm, we conduct extensive experiments on two datasets: the widely used benchmark dataset \uspto ~\cite{Chen2020}, and a new \usptoext~dataset which is ten times the size of \uspto.
On the \uspto~dataset, we achieved a 99.47\% success rate under the constraint of $500$ search steps constraint using single-target search, a new record better than previous results with 2.6 points.
In addition, the plans also has better quality in terms of plan length.
We also achieve a $72.89\%$ success rate on the \usptoext~dataset on a single target search, which is better than baseline systems.
We can future boost the success rate using batch target search.
Finally, our study on the data shows that intra-target and inter-target duplication is critical for this task, and both graph search and GNN guidance contribute to performance improvements.

Our contribution can be summarized as follows:

$\bullet$ We propose a graph search method for retrosynthetic planning task that can eliminate the intra-target duplication in conventional tree search methods.

$\bullet$ We also propose a novel GNN guided policy based on the whole search graph to better select candidate nodes in the search graph.

$\bullet$ The proposed method can naturally search a batch of targets together by eliminating the inter-target redundancy. This can help further improve the success rate.

\section{Preliminaries}

\subsection{Retrosynthetic planning}

Let $\mathcal{M}$ denote the space of all molecules, and let $\mathcal{I}$ denote the collections of available molecules, $\mathcal{I}\subseteq\mathcal{M}$. We have $T$ target molecules to be synthesized, which are denoted as $\mathcal{T} = \{t_i\}_{i=1}^T, t_i \in \mathcal{M}$. The goal is to synthesize all molecules in $\mathcal{T}$ with  the ingredients in $\mathcal{I}$.

Given any $m\in\mathcal{M}$, we might have $n$  ways to synthesize it:
\begin{equation}
R(m)=\{R_1(m), R_2(m),\cdots,R_n(m)\}.
\end{equation}
Each reaction $P_i(m)$ denotes triplet $(m,\mathcal{R}_i(m),c_i(m))$: $m$ is the reaction product, $\mathcal{R}_i(m)\subset\mathcal{M}$ refers to the reactants, and $c_i(m)\in\mathbb{R}_+$ is the reaction cost. If $\exists R_i(m)\in R(m)$ satisfying that  $\forall m^\prime\in\mathcal{R}_i(m)$, $m^\prime\in\mathcal{I}$, then we find a successful path way to synthesize $m$. Otherwise, we need to pick some $R_i(m)$ to further synthesize all the $t^{\prime\prime}\in \mathcal{R}_i(m)\backslash \mathcal{I}$. If the elements in $\mathcal{R}_i(m)\backslash \mathcal{I}$ can be recursively synthesized using one or multiple steps with molecules in $\mathcal{I}$, then we successfully find a path too.

The mapping from $m$ to $P(m)$ is realized by the single-step retrosynthesis model $B$ with parameters $\theta_b$,
\begin{equation}
    B(\cdot; \theta_b) = m \mapsto \{R(m)_i = (m, \mathcal{R}_i, c_i)\}_{i=1}^{K},
\end{equation}
which generate at most $K$ reactions for a given molecule $m$.
Furthermore, there is a limitation of how many times the single-step model $B(\cdot; \theta_b)$ can be used.

\subsection{A* search}
A* search~\cite{hart1968} is an efficient method for many planning tasks and have shown promising results in previous works~\cite{Chen2020, Kim2021}.
In A* search, the total cost of a search state $x$ is defined as
\begin{equation}\label{eq:rawastar}
    f(x) = g(x) + h(x),
\end{equation}
where the total cost $f(x)$ is decomposed into two parts: the known cost $g(x)$ and the $h(v|G)$ the heuristic future cost estimation $h(x)$.
Usually, the $g(x)$ is easy to define because all required information is available and the key challenge is to define an efficient $h(x)$.
\section{PROPOSED METHOD}
\begin{figure*}[!htbp]
    \centering
    \includegraphics[width=\linewidth]{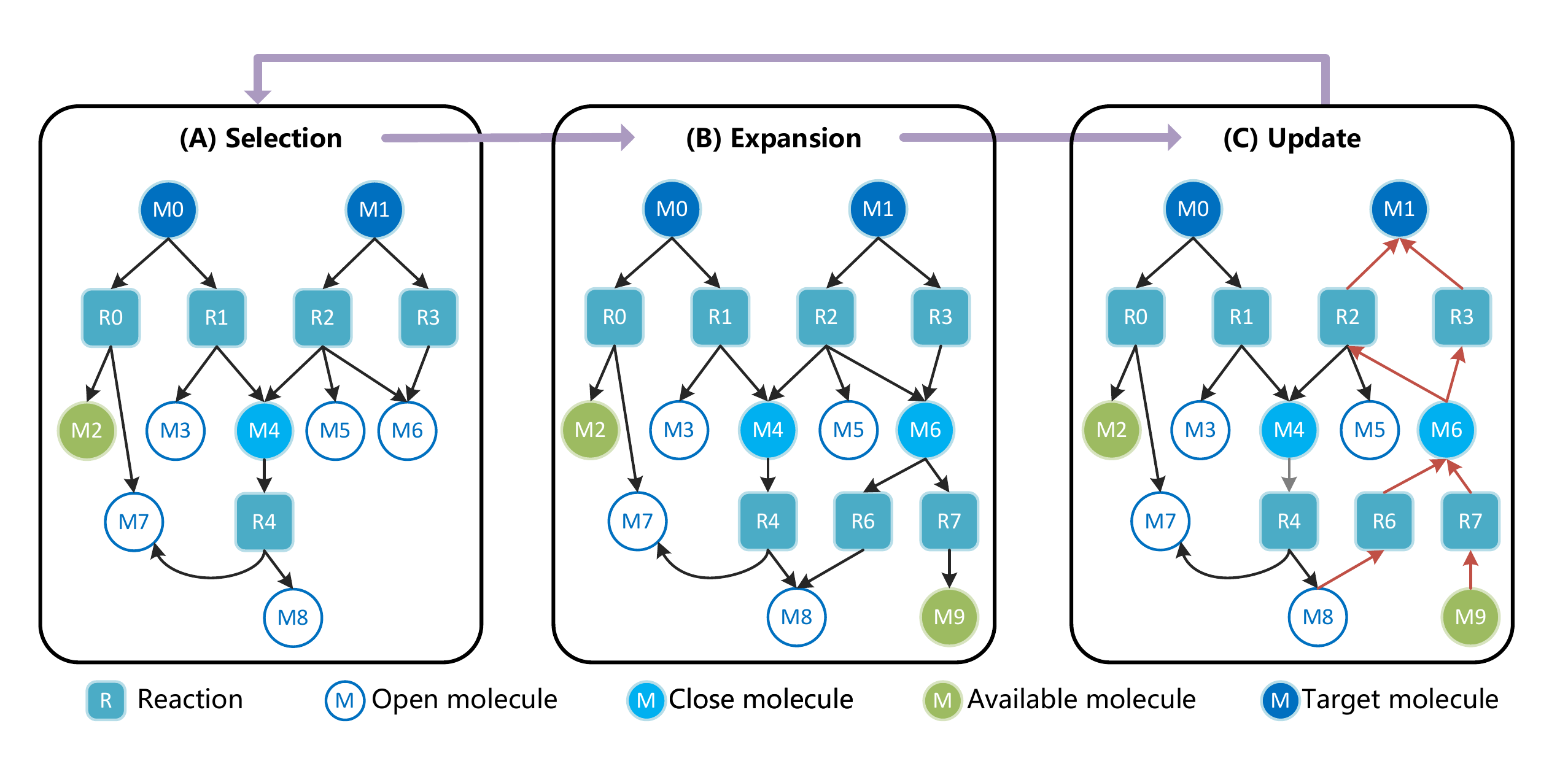}
    \caption{Overview of RetroGraph algorithm. The circles and squares denote molecule and reaction nodes, respectively. The molecule \texttt{M6} is selected for expansion in this round. The dark arrows are synthetic dependencies and the red arrows are bottom-up update process from \texttt{M8} and \texttt{M9}. Note the molecule \texttt{M4}, \texttt{M6}, \texttt{M7}, and \texttt{M8} are shared by multiple reactions.}
    \label{fig:system}
\end{figure*}

\subsection{AND-OR search graph}
Denote the search graph as $G=(\vertset, \edgeset)$, where the $\vertset$ is the node set and the $\edgeset$ is the edge set. $G$ is a directed graph, i.e., given an edge $e = (v_s, v_t)$, $v_s$ is the predecessor and $v_t$ is the successor. 
Different from a tree, each node might have multiple predecessors and successors.

There are two kinds of nodes in the graph $G$: the molecule node and the reaction node. Each node only has one type. 
Let $\vertset_m$ and $\vertset_r$ denote the collections of molecule nodes and reaction nodes. 
We also guarantee that $\vertset_m \cup \vertset_r = \vertset$ and $\vertset_m \cap \vertset_r = \emptyset$.
There is no edge between the nodes of the same type. That is, molecule nodes can only be connected to reaction nodes, and vice versa. In other words, the types of a node and its predecessors/successors are different.  

An example of the search graph is shown in Figure~\ref{fig:system}, where the $M$'s and the $R$'s are molecule nodes and reaction nodes respectively.
The target $M0, M1\in \mathcal{T}$ and the building blocks $M2, M9\in\mathcal{I}$.
$M0$ can be synthesized using reaction $R0$ {\em or} $R1$. If we choose $R0$, we use $M2$ {\em and } $M7$ to synthesize $M0$. If we choose $R1$, we use $M3$ {\em and} $M4$ to synthesize $M0$.
Notably, the nodes $M4, M6, M7, \text{and}, M8$ have more than one predecessors, making the search graph different from a search tree.

Further, the $\vertset_m$ is split into two subsets: the open molecule nodes $\vertset_{\text{mo}}$ and closed molecule nodes $\vertset_{\text{mc}}$.
A molecule node $v \in \vertset_{\text{mc}}$ only if the molecule in available set $\mathcal{I}$ or it has successors.
Otherwise, the node belongs to $\vertset_{\text{mo}}$.
For example, in Figure~\ref{fig:system}(A), $M3,M5,M6,M7,M8 \in \vertset_{\text{mo}}$ and other molecules are closed. 

We define the Boolean function $\success(v) \mapsto \{\text{true}, \text{false}\}$ that evaluate the success state of each node. The search graph satisfy the AND-OR constraint~\citep{Chen2020}.
Mathematically\footnote{To simplify notations, we do not distinguish a node and the molecule/reaction it represents when there is no ambiguity.},
\begin{equation}\label{eq:succ}
\resizebox{0.9\hsize}{!}{%
    $\success(v)=\left\{
    \begin{array}{ll}
        \underset{(v, v_k)\in \edgeset}{\bigwedge}  \left\{ \success(v_k) \right\} & v \in \vertset_r\\
        & \\
         \underset{(v, v_k)\in \edgeset}{\bigvee}  \left\{\success(v_k)   \right\} \vee \{v \in \mathcal{I}\} &  v \in \vertset_m
    \end{array}
    \right.,$
}
\end{equation}
where $\wedge, \vee$ denote logical AND, OR operation, respectively. Specifically,

(1) The reaction nodes are AND nodes. A reaction node is in true status if all its successors are in true status. Intuitively, a successful reaction node means all the reactants for this reaction are obtainable.

(2) The molecule nodes are OR nodes. A molecule node $v$ is in true status if one of the following two constraints are satisfied: (i) the molecule $v$ is in the set $\mathcal{I}$; (ii) it has at least one successor $v_k$, whose status is true. Intuitively, a successful molecule node means that it is already in the available set or we can synthesize $v$ using the ingredients in $\mathcal{I}$.

One difference between a search tree and a search graph is that a graph may contain cycles.
However, cycle avoidance is not easy for two reasons:
First, detecting cycles in a graph is time-consuming because we need to visit all nodes to determine if a graph has a circle or not.
Second, even after a cycle is detected, it is not trivial to decide which nodes and edges to remove to break the cycle.
Fortunately, we notice that cycle will not affect the success checking in Equation~\eqref{eq:succ}.
More specifically, for a node $v \in \mathcal{I}$, the molecule is already available, and no exploration of $v$ is required.
This means the out-degree of $v$ is zero, ensuring that $v$ is never in a cycle.
From the recursive definition in Equation~\eqref{eq:succ}, we can see that all success routes must end at available molecules.
Thus, the cycle dependency will not cause a fake success status.
Therefore, we use a cycle-tolerance strategy in our planning algorithm. 

\subsection{Planning procedure}
Our planning algorithm contains three steps and details are shown in Algorithm~\ref{alg:main}.

\SetKwComment{Comment}{/* }{ */}
\begin{algorithm}[tb]
\caption{RetroGraph Planning}\label{alg:main}
\KwIn{target molecules $\mathcal{T}$, single-step model $B(\cdot;\theta_b)$, available molecules $\mathcal{I}$, planning step budget $N$, single-step model candidate count $K$.}
Initialize search graph $G=(\vertset, \edgeset)$ where $\vertset = \mathcal{T} $ and $\edgeset = \emptyset$\;
\Repeat{$N = 0$ or $\success(t) = \text{true}: \forall  t\in\mathcal{T}$ or $\vert \vertset_{\text{mo}} \vert = 0$ }{
    \tcc{Select the next molecule node $v_\text{next}$ to expand using the Equation~\eqref{eq:gnn_plan}}
    $v_{next} \leftarrow \arg\min_{v\in \vertset_{mo}}{\cost(v|G)}$\;
    
    \tcc{Compute top $K$ reactions by model $B(\cdot;\theta_b)$}
    $\{{R}_i(\vnext)\}_{i=1}^{K} \leftarrow B(v_{\text{next}};\theta_b)$\;
    
    $\Upsilon \leftarrow \emptyset$\;

    \For{$i=1$ {\bfseries to} $K$}{
        Expand $G$ by $\{{R}_i(\vnext)\}_{i=1}^{K}$ using Equation~(\ref{eq:u1},~\ref{eq:u2})\;
        Update the affected nodes to $\Upsilon$\;
    }
    
    \For{$v \in \Upsilon$}{
        Bottom-up update status of $G$ from $v$\;
    }
    $N \leftarrow N-1$\;
}
\end{algorithm}

\paragraph{Step A: Selection}
Given a search graph $G$, the next node to further explore is
\begin{equation}
v_{next}  = \mathop{\arg\min}_{v\in \vertset_{mo}}{\cost(v|G)},
\end{equation}
where $\cost$ refers to the estimated cost of exploring $v$ conditioned on the existing graph $G$. A good $\cost$ should consider both existing cost of obtaining $v_{\rm next}$ and its future cost. We will introduce more details about $\cost$ is Section~\ref{sec:gnn}.

\paragraph{Step B: Expansion}
This step expand $G$ with $\vnext$.
Following~\cite{Chen2020,Kim2021}, we first extract the molecule features (i.e., Morgan fingerprint) of $\vnext$, and then feed it into the single-step model $B(:, \theta_b)$.
The single-step model then predicts the top $K$ reactions\footnote{More specifically, it predicts the top $K$ reaction templates. All templates are validated by Rdchiral~\citep{coley2019rdchiral}, a RDKit wrapper,  to remove the chemical impossible ones.}. These reactions are then applied to $\vnext$, and we obtain the reactants and the corresponding costs.
Finally, we update the search graph $G(\vertset, \edgeset)$ with reactions $\{{R}_i = (\vnext, \mathcal{R}(\vnext)_i, c(\vnext)_i)\}_{i=1}^{K}$ 
\begin{align}
    \vertset & \leftarrow \vertset \cup \{m_{i,j} | 1 \le i \le K, m_j \in \mathcal{R}_i(\vnext) \}, \label{eq:u1} \\
    \edgeset & \leftarrow \edgeset \cup \{ e(v_{\vnext}, R_i) \}_{i=1}^K \nonumber \label{eq:u2}\\
    & \quad \cup \{ e(R_i, m_j) | 1 \le i \le K, m_j \in \mathcal{R}_i(\vnext) \}.
\end{align}

An important property of this method is that no duplicated molecules are allowed in the graph.
Therefore, we leverage a global molecule memory to store all molecules that already exists in the graph and void creating new one when it is already exits.

\paragraph{Step C: Update}
After graph expansion,we need update the status (e.g., success state, historical cost) of all nodes.
Here we use a bottom-up strategy and only update the affected nodes to save search time.
More specifically, Let $\Upsilon$ denotes the nodes need to be updated, which is initialized by the molecules directly affected in Step B.
For each molecule node $v \in \Upsilon$, we find all edges $e(v_p, v)$ that ends at $v$.
Then for each $v_p$, we recompute success state and historical cost.
We keep this process until all required nodes are updated.

We repeat these three steps until the termination condition is stratified.
i.e, we have successfully found routes for all targets, used up all iteration budgets, or there are no nodes to expand.
After this search process, we can extract synthetic routes as tree from the search graph because the tree structure synthetic routes are more accessible to humans.
Therefore, we apply a depth-first iteration strategy on a success search graph.
To be more specific, we start from a target molecule and select all success reactions of it. Then, we recursively visit the reactants until we meet the leaf nodes.
We can build the synthetic tree from bottom to top during the recursive visit process.

\subsection{GNN guided policy}\label{sec:gnn}
For the retrosynthetic planning, we can write the A* cost function in Equation~\eqref{eq:rawastar} as
\begin{equation}\label{eq:astar}
    \cost(v|G) = g(v|G) + h(v|G),
\end{equation}
where the $g(v|G)$ is the known (historical) cost from the start point (a target node $t$ in this task) to $v$ and the $h(v|G)$ is the heuristic function that estimate the future cost.

Let $\Psi(v_1, v_2, \cdots, v_n)$ denotes a path of graph $G(\vertset, \edgeset)$ where $v_i \in \vertset: \forall v_i \in \Psi$ and $e(v_i, v_{i+1}) \in \edgeset: \forall  v_i \in \Psi (i<n)$.
The history cost is defined as 
\begin{equation}\label{eq:costh}
    g(v|G) = \min_{t\in\mathcal{T}} \min_{\Psi(v_t,\cdots,v)} \sum_{v_i\in \Psi \cap \vertset_r} c_i(v),
\end{equation}
This means we select a target $t\in\mathcal{T}$ and a path $\Psi(t,\cdots,v)$ from $t$ to $v$ to compute a lower-bound of the history cost.

\begin{figure}[!htbp]
    \centering
    \includegraphics[width=\linewidth]{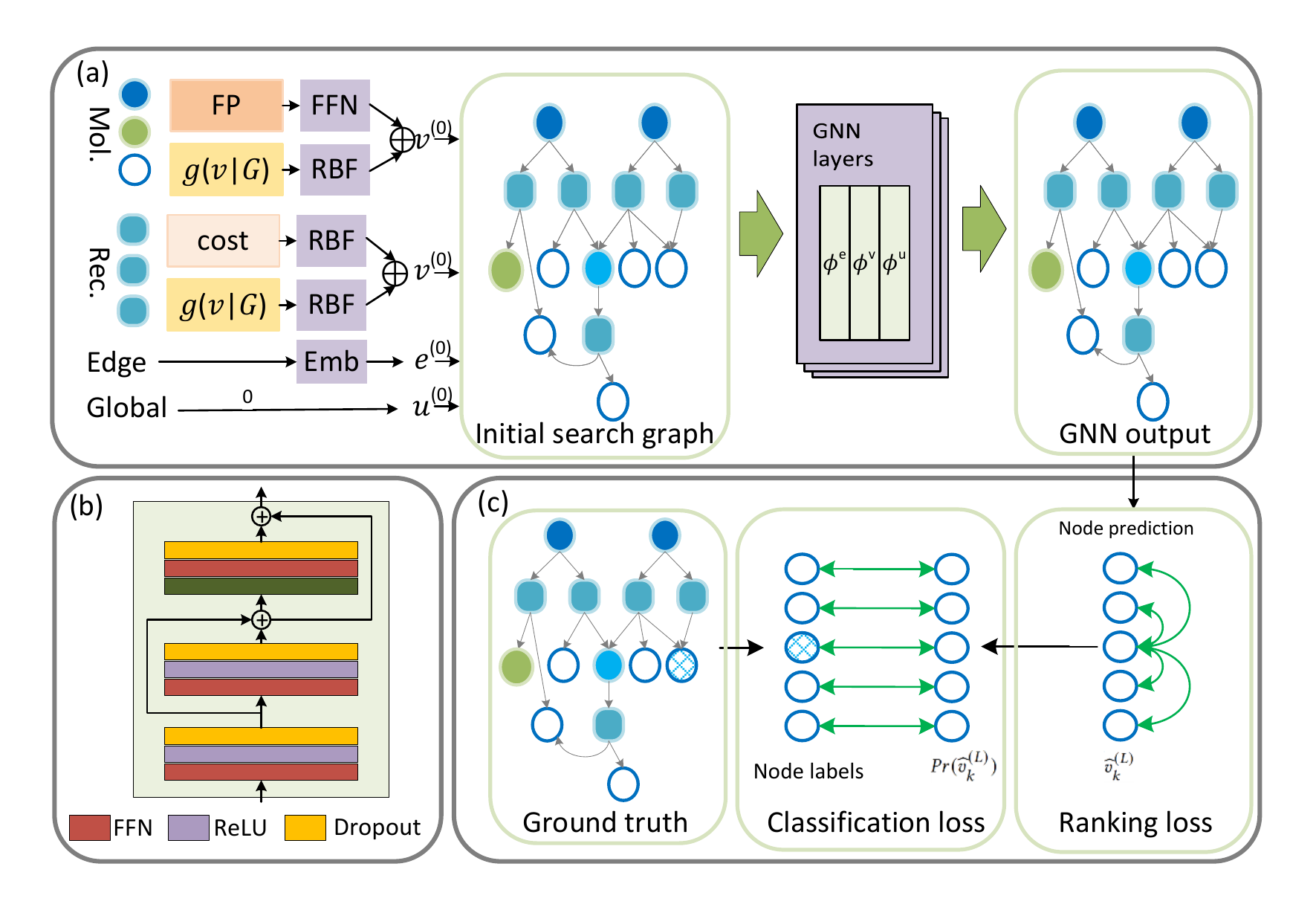}
    \caption{Illustration of policy GNN. (a) Graph representation and GNN process; (b) Illustration of MLP in Equation~(\ref{eq:mlp1},\ref{eq:mlp2},\ref{eq:mlp3}); (c) Computation of classification loss and ranking loss.}
    \label{fig:gnn}
\end{figure}

However, defining the $h(v|G)$ on an AND-OR search graph is more complicated.
Because of the reaction nodes (AND nodes), the total route cost of a given molecule is related to itself and other molecules in the route.
Therefore, complex computation in $G$ is required to if we want to have a high-quality estimation of $h(v|G)$.
Instead, we propose a learning-based method and use a GNN to guide the search.
More specifically, we modify the Equation~\eqref{eq:astar} from the original form to 
\begin{equation}\label{eq:gnn_plan}
    \cost(v|G) = g(v|G) + \text{GNN}(v|G; \theta_{G}),
\end{equation}
where the $\text{GNN}(v|G; \theta_{G})$ is a GNN with parameter $\theta_{G}$ to guide the search.
The GNN architecture is in Figure~\ref{fig:gnn}.

\noindent\textbf{GNN architecture.}
Our GNN consists three components: node and edge embeddings, GNN layers, and output layer.
First, we embed the node and edge features into the hidden representations.
For molecule nodes, we use the historical cost and Morgan Fingerprint (FP) as the node feature, and for reaction nodes, we use the historical cost and reaction cost as node feature.
Meanwhile, we use different edge embedding for different directions of edges as edge feature.
Finally, we use a zero vector to initialize the global (i.e., graph level) representation.
The FP is project to the hidden dimension using a feedforwad (FFN) layer.
All scale values (i.e., historical cost, reaction cost) is embedded by the Radial Basis Function (RBF) kernel, which is defined as:

\begin{equation}
    \text{RBF}(x;L,H,N)_i = \exp\left(
    \vcenter{\hbox{$\displaystyle
    {\frac{-\left(x-i*\frac{H-L}{N}\right)^2}{\tau}}
    $}}
    \right), \, 0 \le i < N,
\end{equation}
where $L, H, N$ are the lowerbound, upperbond, and embedding size, respectively.
Intuitively, this function embed a number $x\in\mathbb{R}$ using a series of Gaussian distributions.
If we denote the node, edge, and global representation at $i-$th layer $v^{(i)}, e^{(i)}, u^{(i)}$, respectively, the initialization process can be written as:
\begin{align}
    e^{(0)}_k &= \text{embed}(\text{dir}(e_k)), \\
    v^{(0)}_k &= \text{RBF}\left(\text{cost}_h(v)\right) \oplus ( \text{FFN}(\text{FP}(v_k)): \; v_k \in \vertset_m,  \\
    v^{(0)}_k &= \text{RBF}\left(\text{cost}_h(v)\right) \oplus ( \text{RBF}(c_k)) : \; v_k \in \vertset_r, \\
    u^{(0)} &= \vb{0}.
\end{align}

To model both the global and local structure of the search graph, we use meta GNN Layer~\cite{battaglia2018relational} as the building layers of the policy GNN.
The meta GNN layer consists of three sub-layers that update the edge, node, and global representations.
Suppose we have $L$ GNN layers, for $1 \le i \le L$, the update process is
\begin{align}
        e^{(i)}_k &= \phi^e\left(e^{(i-1)}_k, v^{(i-1)}_{sk}, v^{(i-1)}_{kt}, u^{(i-1)}\right), \\
        v^{(i)}_k &= \phi^v\left(\mathrm{e}^{(i)}_k, v^{(i-1)}_k, u^{(i-1)}\right), \\
        u^{(i)} &= \phi^{u}\left(e^{(i)}, v^{(i)}, u^{(i-1)}\right).
\end{align}
Where $v_{\text{sk}}, v_{\text{kt}}$ means the start and end node of edge $e_k$ and $\mathrm{e}_k$ denotes all edges $e$ end at $v_k$.
In our method, we use:

\begin{align}
    \phi^e &= \text{MLP}\left(e^{(i-1)}_k \oplus v^{(i-1)}_{sk} \oplus v^{(i-1)}_{kt} \oplus u^{(i-1)}\right), \label{eq:mlp1}\\
    \phi^v &= \text{MLP}\left( v^{(i-1)}_k \oplus \text{msg}^{(i)}(v_k) \oplus u^{(i-1)} \right), \label{eq:mlp2}\\
    \phi^{u} &= \text{MLP}\left(u^{(i-1)} \oplus \frac{1}{N}\sum{v^{(i)}_k} \right), \label{eq:mlp3}
\end{align}
where
\begin{equation}
    \text{msg}^{(i)}(v_k) = \frac{1}{|\mathrm{e_k}|}\sum_{e_{sk}=(v_s, v_k) \in \mathrm{e_k}} \text{MLP}\left(v^{(i-1)}_s \oplus e^{(i-1)}_{sk}\right).
\end{equation}
The $\oplus$ denotes the tensor concatenation operation and $\text{MLP}$ denotes three-layer perception networks using ReLU activation~\cite{agarap2018deep}, residual connection~\cite{he2016}, and dropout~\cite{Srivastava2014} and the details are in Figure~\ref{fig:gnn}.

Finally, we use a feedforwad layer on each node to transform the node hidden representation in the last GNN layer $v_k^{(L)}$ to a scalar value $\widehat{v}^{(L)}_k$ and use the sigmoid function to convert $v_k$ to probability $Pr(\widehat{v}^{(L)}_k)$ for classification loss.
During search time, we normalize all open nodes prediction to get the score for selection,
\begin{equation}
    \text{GNN}(v_k|G;\theta_G) = \frac{\exp\left(\widehat{v}_k^{(L)}\right)}{\sum_{v\in\vertset_{mo}}\exp\left(\widehat{v}^{(L)}\right)}.
\end{equation}

\noindent\textbf{GNN training}
We use an offline-training strategy to train the GNN policy network.
More specifically, we first use A* search on the training dataset to find synthetic plans for all target molecules.
Next, we gradually follow the generated reactions for each successful plan to expand the search graph.
We collect the graph representations as GNN input and the correct nodes to expand as GNN labels for each expansion step.  
We mark the correct nodes for expanding with positive labels and other open nodes with negative labels.

We treat the GNN training as both node classification and ranking tasks.
First, we have a binary classification cross-entropy (CE) loss for each node because there could be more than one positive label.
The loss function is defined as

\begin{equation}
    \mathcal{L}_{bce} = - \frac{1}{|\vertset_{\text{mo}}|}\sum_{v\in\vertset_{\text{mo}}} y \log(Pr(\widehat{v})) - (1-y)\log(1-Pr(\widehat{v})).
\end{equation}
Following~\citet{Chen2020}, we also use a ranking loss that the $\widehat{v}$ of positive nodes are larger than the negative nodes with at least $\tau$ margin.
\begin{equation}
    \mathcal{L}_{rank} = -\frac{1}{|\vertset_{op}|}\frac{1}{|\vertset_{on}|}\sum_{v_p \in \vertset_{op} }\sum_{v_n \in \vertset_{on}} \min(0, \widehat{v}_p - \widehat{v}_n - \tau).
\end{equation}
where $\vertset_{op}, \vertset_{on}$ are positive and negative open nodes set, respectively.
Therefore, the final loss of GNN is 
\begin{equation}
    \mathcal{L}_{\text{GNN}} = \mathcal{L}_{bce} + \mathcal{L}_{rank}.
\end{equation}
In all our experiments, we use $\tau = 4$ and do not tune the weights of these two loss functions.

\section{Experiments}
We present experimental results to answer the following questions:

\noindent\textit{Q1 (Section~\ref{sec:res_single}):} How does our algorithm perform in single target search when eliminating intra-target redundancy?

\noindent\textit{Q2 (Section~\ref{sec:sec:res_batch}):} How does our algorithm perform in batched target search when eliminating inter- and intra-target redundancy?

\noindent\textit{Q3 (Section~\ref{sec:res_study}):} How frequently do inter-target and intra-target occur in tree search?

\noindent\textit{Q4 (Section~\ref{sec:res_ablation}):} 
How do the graph search and the GNN policy network contribute to our system?

\subsection{Experimental Setup}\label{sec:exp_setup}

\textbf{Dataset.}
We use two datasets to evaluate our method. The first one is a widely used benchmark \uspto~dataset that was introduced by Retro*~\citep{Chen2020}.
Considering that the scale of test data is small, we created an extra test set named \usptoext~that contains $10 \times$ sizes of test routes of \uspto.
For single-step model training, we use the $1.3M$ reactions from~\citet{Chen2020} and follow~\citep{Kim2021} to create augmented training data using forward and backward models.
For GNN training data, we follow the steps of Section~\ref{sec:gnn} and collected about $140k$ graphs.
We then randomly select $1,000$ graphs for the validation and $1,000$ graphs for the test and use the remaining graphs for training.
Meanwhile, we use the same set of commercially available molecules set as~\cite{Chen2020,Kim2021}.
More details are in the Appendix.

\noindent\textbf{Single-step model.}
Although our planning framework is generally compatible with any single-step model,
we use the same model as~\citet{Chen2020, Kim2021} to make a fair comparison with previous work.
It is a template-based single-step retrosynthesis model that is introduced by~\citet{segler2017towards},
We use the same reaction template set as~\citet{Chen2020}, which contains about $380k$ templates extracted from the original USPTO dataset using Rdchiral~\citep{coley2019rdchiral}.
Meanwhile, the model architecture is a 2-layer MLP using the Morgan fingerprint~\citep{rogers2010extended} as input and the probability of using each template as output.
It means it treats the single-step retrosynthetic as a multi-class classification problem.
The fingerprints are extracted with radius $2$ and $2048$ bits.
We use the top 50 predicted templates to expand the input molecule during planning.
We use the same hyper-parameters as~\citet{Kim2021} to train the MLP model, with a learning rate of $0.001$, dropout rate $0.4$, and batch size $1024$.
The models are trained for 20 epochs using Adam optimizer~\citep{kingma2014adam}.

\noindent\textbf{Policy GNN.}
We use three Meta GNN layers~\cite{battaglia2018relational} as the building blocks of policy GNN with hidden dimensions as $128$.
For RBF kernel, we use $L=0, H=10, N=64, \text{and}\, \tau=\frac{(H-L)^2}{4}$ in all our experiments.
The model is trained with Adam optimizer~\citep{kingma2014adam} using a learning rate of $0.0001$ and dropout $0.1$.
We use $32$ graphs per mini-batch and train the network with a max epoch of $20$.
The best checkpoint is selected by the ranking loss on the validation set.

\noindent\textbf{Evaluation.}
We use \uspto~and \usptoext~datasets to evaluate our method and compare it with previous works.
More specifically, we measure the success rate with different iteration limits.
For \uspto, we follow previous works to use $500$ as the max iteration limit, and for \usptoext, we use $100$ to make the task more challenging.
In addition to the success rate, we also show the results of average iteration, the average number for molecule nodes, the average number of reaction nodes under the limit of max iterations.
Finally, we also compute the average route length and average route cost to evaluate the plan quality.

\noindent\textbf{Baselines.}
We compare our system with representative baselines on this task and the details are available in Appendix.
Briefly, \textbf{Greedy DFS}~\citep{Hong2021} is a classic planning method that always prioritizes the node with max probability; 
\textbf{DFPN-E}~\citep{kishimoto2019depth} is a Depth-First Proof-Number search method; 
\textbf{MCTS-rollout}~\citep{Hong2021}, \textbf{EG-MCTS}~\citep{Hong2021}, and  \textbf{EG-MCTS-0}~\citep{Hong2021} are MCTS methods;
\textbf{Retro*}~\citep{Chen2020}, \textbf{Retro*-0}~\citep{Chen2020}, \textbf{Retro*+}~\cite{Kim2021}, and \textbf{Retro*+-0}~\cite{Kim2021} are A* search methods.

\begin{table*}[!htbp]
\centering
\caption{Experimental results on USPTO dataset with single target search. We compare each algorithm at the success rate of different limit. Under the limit of 500, we also show the average number of iterations , reaction (Rec.) nodes, and molecule (Mol.) nodes. The best results are marked as bold.}
\label{tab:main_single}
\begin{tabular}{lcccccSSS}
\toprule
\multirow{2}{*}{\textbf{Algorithm}} & \multicolumn{5}{c}{\B Success Rate of Iteration Limit {[}\%{]} $\uparrow$} & {\B\multirow{2}{*}{ \# Iteration $\downarrow$}} & {\B \multirow{2}{*}{\# Rec. Nodes$\downarrow$}} & {\B\multirow{2}{*}{\#  Mol. Nodes$\downarrow$}} \\ \cline{2-6}
                           & 100       & 200       & 300       & 400      & 500      &           &              &              \\ \midrule
Greedy DFS       &  38.42    &  40.53    &  44.21  &    45.26 &     46.84&     300.56&     {-}        &     {-}        \\ 
DFPN-E & 50.53 & 58.42 & 64.21 & 68.42 & 75.26 &208.12 & 3123.33 & 4635.08 \\
MCTS-rollout&43.68&47.37&54.74&58.95&62.63& 254.32 & {-} & {-} \\
Retro*-0& 36.84& 59.47&68.95 & 74.74& 79.47&210.49 & 3908.15& 5575.97\\
Retro*& 52.11&66.32 &76.84 &81.05 & 86.84 & 166.72& 2927.92& 4174.52 \\
Retro*+-0&67.37&82.10&93.16&95.26&96.32& 96.14& 1421.90& 2108.50 \\
Retro*+&71.05&85.26&88.95&90.00&91.05& 100.15& 1209.79& 1767.81 \\
EG-MCTS-0&57.37&63.68&68.42&71.05&73.68&186.15& 2525.20& 3339.52 \\ 
EG-MCTS&85.79&92.63&94.21&95.79&96.84& 55.84& 869.59& 1193.79\\
\midrule
\textbf{RetroGrph (Ours)} & \textbf{88.42} & \textbf{97.89} & \textbf{98.95} & \textbf{99.47} & \textbf{99.47} & \textbf{45.13} &\B 674.22 &\B 500.43 \\
\bottomrule
\end{tabular}
\end{table*}

\begin{table*}[!htbp]
\centering
\caption{Experimental results on USPTO-EXT dataset with single target search. We compare each algorithm at the success rate of different iteration limit. Under the limit of 100, we also show the average number of iterations, reaction (Rec.) nodes, and molecule (Mol.) nodes. The best results are marked as bold.}
\label{tab:main_single_ext}
\begin{tabular}{lccccccSSS}
\toprule
\multirow{2}{*}{\B Algorithm} & \multicolumn{6}{c}{\B Success Rate of Iteration Limit {[}\%{]} $\uparrow$} & {\multirow{2}{*}{\B \# Iteration $\downarrow$}} & {\B \multirow{2}{*}{\# Rec. Nodes$\downarrow$}} & {\B \multirow{2}{*}{\#  Mol. Nodes$\downarrow$}} \\ \cline{2-7}
                        &   10 & 20       & 30       & 40       & 50      & 100      &           &              &              \\ \midrule 
Retro*-0& 42.11& 48.11& 50.63&52.16 & 53.11& 56.68&49.81 &836.54 &1198.50 \\
Retro* & 42.47& 48.79 & 51.84 & 53.63 & 55.00 & 57.89 & 48.49 & 790.49 & 1136.51 \\
Retro*+-0 &48.74 &54.11 & 57.21& 58.68& 60.42&66.16 &42.55 & 535.04& 806.52\\ 
Retro*+ & 49.05& 55.00& 59.11& 61.42& 63.63& 68.74&40.11 &469.42 &716.10 \\
\midrule
\textbf{RetroGraph (Ours)} & \textbf{50.84} & \textbf{58.05}& \textbf{62.05}& \textbf{64.26}& \textbf{66.89}& \textbf{72.89}& \textbf{37.25} &\B  491.97 &\B  373.80\\
\bottomrule
\end{tabular}

\end{table*}

\subsection{Single target planning}\label{sec:res_single}

When the $|\mathcal{T}| = 1$, our method can search one target at a time.
Therefore, we can make a fair comparison with the existing single target planning method.
The results for \uspto~and \usptoext~are shown in Table~\ref{tab:main_single} and Table~\ref{tab:main_single_ext}, respectively.
Under all iteration limits, the success rates of our method are much better than baseline results.
More specifically, when the iteration limit is $500$, we achieved a $99.47\%$ success rate, which is $2.63$-points better than the previous state-of-the-art success rate.
Meanwhile, the average iteration of our method is only $45.13$, which is $10.7$-point lower than \texttt{EG-MCTS}.
For the average number of reaction and molecule nodes, we achieved $674.22$ and $500.43$, respectively.
The average number of molecule nodes is only about half of the previous results because redundant molecule nodes are eliminated.
These numbers demonstrate the correctness of our motivation and effectiveness of this method.
In Table~\ref{tab:single_quality}, we show the result of plan quality.
Compared with baselines, we achieved lower route length and cost.
Our average route length is $6.33$, better than previous best results.
Meanwhile, the route cost is better than most baselines and comparable with the best result.
These results exhibit the effectiveness of our method.

\begin{table}[!htbp]
    \centering
    \caption{Experimental results of plan quality on USPTO.}
    \label{tab:single_quality}
    \begin{tabular}{lSc}
        \toprule
        \textbf{Algorithm} & {\textbf{Route Length}} & \textbf{Route Cost} \\ \hline
        Retro*-0 & 11.21& 19.40\\
        Retro* & 9.71 & 15.33 \\
        Retro*+-0 & 7.69 & \textbf{11.66} \\
        Retro*+ & 8.74 & 15.23\\
        \hline
        \textbf{RetroGraph (Ours)} &\B 6.33 & 12.92  \\
        \bottomrule    
    \end{tabular}

\end{table}

\subsection{Batch targets planning}\label{sec:sec:res_batch}

\begin{figure}[!tb]
    \centering
    \includegraphics[width=\linewidth]{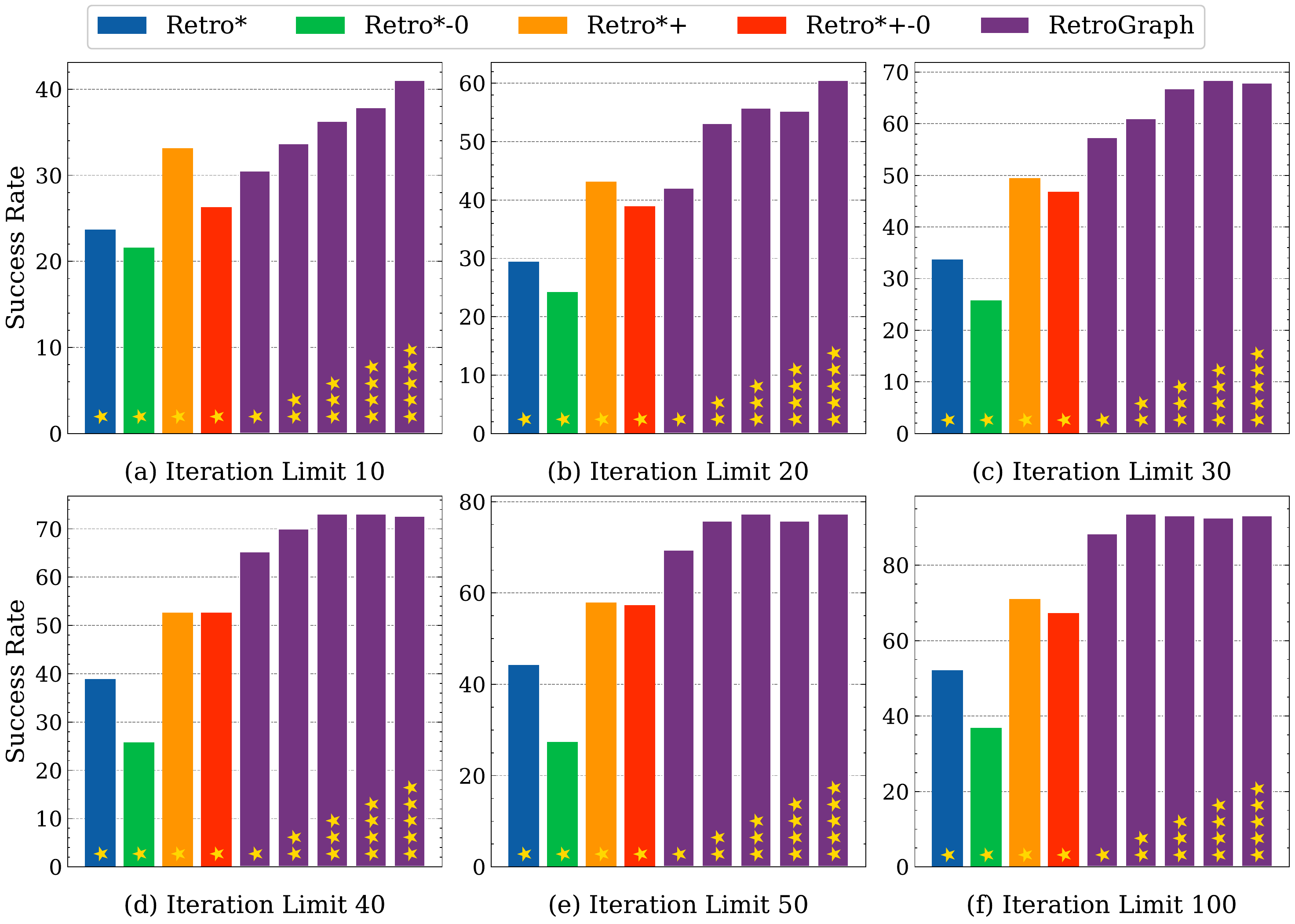}
    \caption{Batch targets planning on USPTO dataset. The number of starts in a bar denotes the number of targets.}
    \label{fig:batch_190}
\end{figure}

One advantage of our method is that we can search a batch of target molecules together, which is not supported by tree search because no nodes are shared.
To evaluate the effectiveness of batch targets planning, on \uspto~, we use K-Means~\citep{lloyd1982} with Morgan fingerprint to cluster all test molecules into $32$ clusters and make mini-batches inside each cluster with different batch sizes.
The results are in Figure~\ref{fig:batch_190} where each sub-figure represent different iteration limits.
Meanwhile, the number of stars inside the bar represents the number of targets in a batch
To make the problem more challenging, we reduce the max iteration budget to $100$.
From the results, we have the following findings:

\noindent$\bullet$ Under all iteration limits, our method (purple bars), regardless of using single target or batch targets planning, has better performance than conventional tree-based methods.

\noindent$\bullet$ Under all iteration limits, the batch targets planning have better performance than all tree-based and graph-based single-target search baselines.

\noindent$\bullet$ With the increase of batch size, the performance will improve, but the performance will be saturated. It will be interesting to study the larger batch in future work.

\subsection{Study of duplication in tree search}\label{sec:res_study}
\begin{figure}[!htbp]
    \centering
    \includegraphics[width=0.9\linewidth]{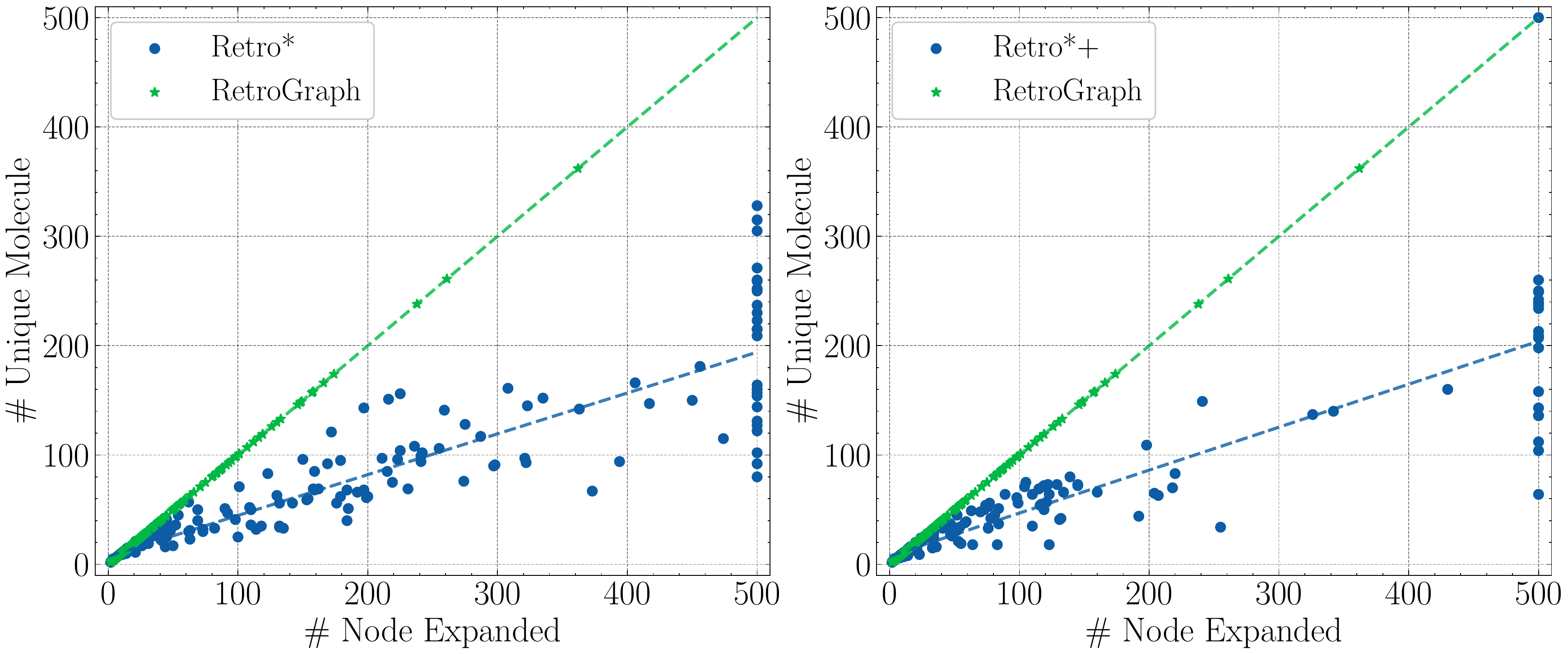}
    \caption{Study of intra-duplication in tree search. The left and right plots are results of Retro* and Retro*+, respectively.}
    \label{fig:dup}
\end{figure}

To study the intra-target redundancy, we use tree-based Retro* and Retro*+ on the \uspto~test data and the results are shown in Figure~\ref{fig:dup}.
The horizontal axis is the total number of expanded nodes and the vertical  axis is the number of unique molecules visited.
The dash lines are the linear regression of data of each model.
We can see that the regression line is much lower than the diagonal line, showing that intra-target redundancy is prevalent in the data.
On the contrary, our graph search method does not have redundancy, and the data points will lie on the diagonal line.

To study the inter-target redundancy, we take the synthetic plans in the \uspto~training set as oracle plans.
Then we collect all intermediate reactants in oracle plans and count the number of times each reactant occurs in the plans. 
We observe that the distribution is long-tail. The most popular reactant \textit{acetyl acetate} occurs in $2,989$ routes, which is nearly $1\%$ of all oracle routes.
Furthermore, on average, each reactant occurs $\mathbf{1.77}$ times in the routes, showing that inter-target redundancy is common.

The above analysis shows that intra-target and inter-target duplication are common issues in this task.
Solving this issue is critical for performance.

\subsection{Ablation study}\label{sec:res_ablation}
\begin{figure}[!htbp]
    \centering
    \includegraphics[width=0.8\linewidth]{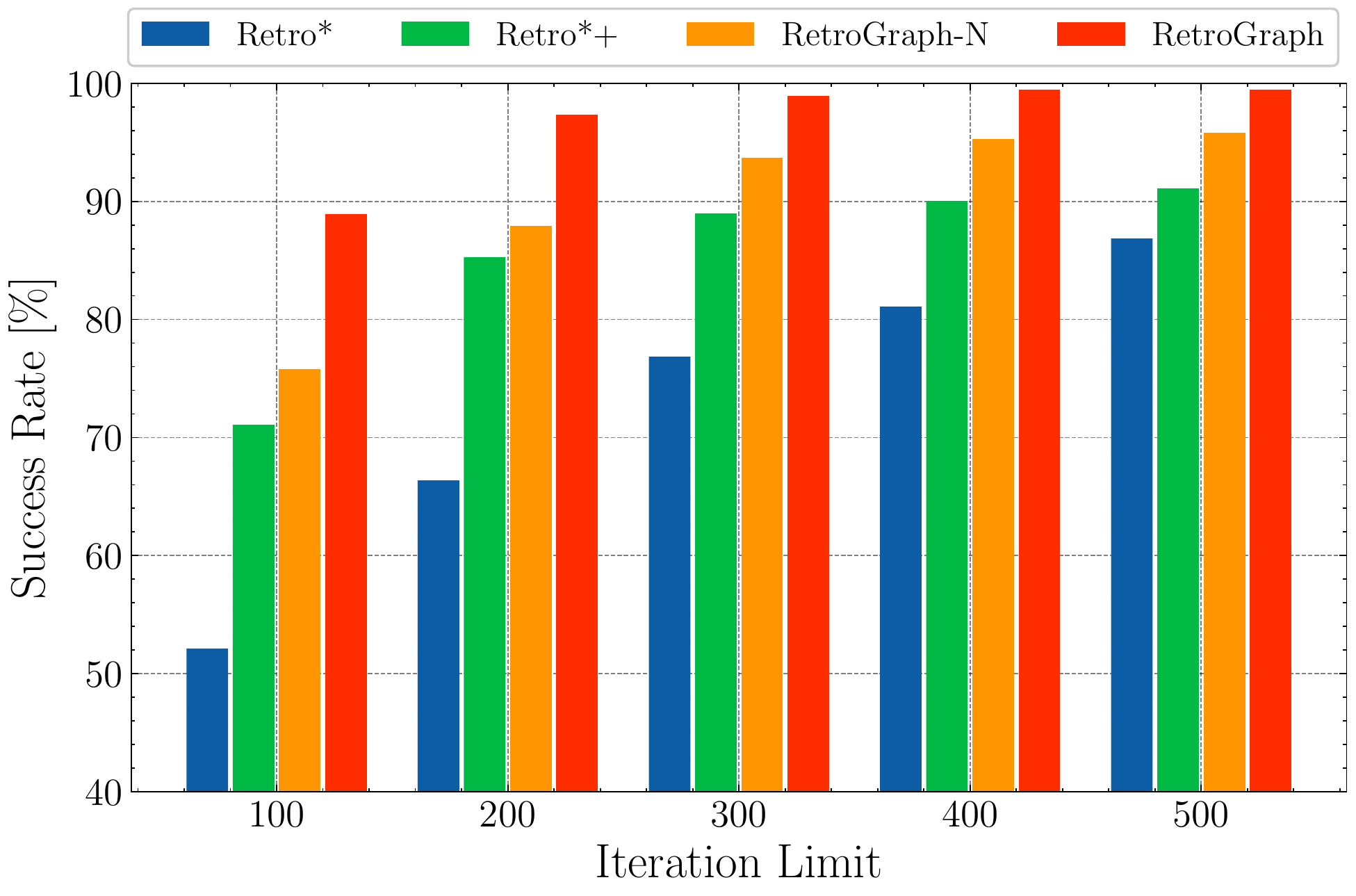}
    \caption{Ablation study on the USPTO dataset.}
    \label{fig:ablation}
\end{figure}

We conducted an ablation study on the \uspto~dataset, and the results is available in Figure~\ref{fig:ablation} where the x-axis denotes the iteration limit, and the y-axis denotes the success rate.
To study the effect of our GNN policy network, we keep the graph search structure and replace the GNN with the value network used by Retro* and Retro*+.
The value network is only based on single-molecule fingerprints and have no access to the complete search graph.
We denote this system as RetroGraph-N.
We also use slashes to mark the background of tree-based methods.
From the results, we can observe that:

\noindent$\bullet$ Comparing Retro*, Retro*+, and RetroGraph-N, the three models that use identical value model checkpoint, the RetroGraph-N has the best performance, indicating that graph search is critical. 

\noindent$\bullet$ The graph-based search methods (i.e., the RetroGraph and the RetroGraph-N) perform better than the tree-based methods (i.e., Retro* and  Retro*+), demonstrating that graph search is effective.

\noindent$\bullet$ Among graph-based methods, the RetroGraph has better performance than RetroGraph-N, showing that guiding the search by GNN is also essential for this task.

\section{Related Work}
\subsection{Retrosynthetic planning}
The data-driven retrosynthetic planning have attracted many research attention ~\cite{Chen2020,Kim2021,Dong2021,Finnigan2021,Jeong2021,segler2018planning,Schreck2019,Wang2020,Hong2021,han2022gnn}.
\citet{Chen2020} propose the Retro* algorithm, who solve this problem with neural guided A* search.
Then \citet{Kim2021} future improve this method with self-training and forward models.
Meanwhile, the Monte Carlo tree search (MCTS) based methods have also been applied to this task~\citep{segler2018planning,Schreck2019,Wang2020,Hong2021}.
Despite their achievements, all of these works are based on tree search and inevitably suffer from intra-target duplication.
Meanwhile, there are no special designs for inter-target duplication in these methods, and they handle each target separately.
The key difference is that our method is a graph-based method without duplicated nodes in trees.
Furthermore, our method is optimized for searching a batch of targets together.

\subsection{Single-step retrosynthesis}
In recent years, many deep learning based single-step retrosynthesis method have been proposed with promising results~\citep{segler2017towards,dai2020retrosynthesis,Chen2021,yan2020retroxpert,somnath2020learning,shi2020graph,yang2021}.
Nevertheless, the goal of single-step retrosynthesis is only to predict reactions of one step, not considering the whole retrosynthetic route.
This is the crucial difference between single-step retrosynthesis and retrosynthetic planning.
The single-step retrosynthesis models are critical component of retrosynthetic planning.
Therefore, we use the same single-step model architecture as baseline systems.

\subsection{Reinforcement learning}
Retrosynthetic planning can be seen as a single player game where the agent (i.e., deep learning models or chemists) manipulates the molecules to a successful state where all ingredients are available.
Reinforcement learning based methods have achieved significant results on many conventional games such as GO~\citep{silver2017mastering,alphazero}, chess~\citep{alphazero}, poker~\citep{poker2019}, and mahjong~\citep{li2020suphx}.
However, retrosynthetic planning is different from those conventional games because of the AND-OR relation and inter-target and intra-target duplication.
Therefore, we propose to use the AND-OR graph with GNN guidance to handle the search progress.

\section{Conclusion}
This work proposes a novel graph-based search algorithm for the retrosynthetic planning task and a GNN guided search policy. Compared with the conventional tree-based methods, our algorithm can reduce the duplication of molecules in the search process. Furthermore, the GNN policy can better handle the complexity of the AND-OR structure of the search graph and suggest that the node expand more effectively. Therefore, our method is more efficient in single-target planning. Furthermore, our method can naturally search a batch of targets and eliminate inter-target duplication. Our method gets a higher performance over previous systems on two benchmark datasets. More specifically, we achieved a 99.45\% success rate with 500 steps limit in the \uspto~dataset, outperforming previous state-of-the-art with 2.6 points.

\begin{acks}
This work was supported by National Natural Science Foundation of China (NSFC Grant No.~62122089 and No.~61876196), Beijing Outstanding Young Scientist Program NO. BJJWZYJH012019100020098, and Intelligent Social Governance Platform, Major Innovation \& Planning Interdisciplinary Platform for the ``Double-First Class'' Initiative, Renmin University of China.
This work was also supported in part by Independent Research Fund Denmark under agreement  8048-00038B.
We wish to acknowledge the support provided and contribution made by Public Policy and Decision-making Research Lab of RUC.
Rui Yan is supported by Beijing Academy of Artificial Intelligence (BAAI).
\end{acks}
\bibliographystyle{ACM-Reference-Format}
\bibliography{sample-base}

\appendix
\section{Data details}
We use two datasets to evaluate our method. The first one is a widely used benchmark \uspto~dataset that was introduced by Retro*~\citep{Chen2020}. The dataset is extracted from the United States Patent Office data that contains about $3.8M$ chemical reactions.
\citet{Chen2020} processed this data and collected $299,202$ training routes, $65,274$ validation routes, and $190$ test routes for retrosynthetic planning task.
They select the $190$ test routes to guarantee the difficulty of evaluating the algorithm capacity.
However, the scale of test data is still small.
Therefore, we created an extra test set named \usptoext~that contains $10 \times$ sizes of test routes of \uspto.
To achieve this goal, we first collect $10M$ molecules from PubChem~\citep{pubchem} and remove those already in training, validation, or test set of \uspto.
Next, we compute the Levenshtein distance~\citep{levenshtein1966binary} between molecules in PubChem and molecules in \uspto.
For each target in \uspto~test set, we keep the top $10$ most similar molecules from PubChem for the \usptoext~testset.
Therefore, we finally collected $1,900$ test targets in \usptoext.
For single-step model training, \citet{Chen2020} also provided about $1.3M$ reactions.
In addition, we follow \citep{Kim2021} to create augmented training data using forward and backward models.
We use the same forward (synthetic) model as \citet{Kim2021} and train our backward (retrosynthetic) models. 
We also use the same data filter threshold $\epsilon = 0.8$ as \citet{Kim2021}.

For GNN training data, we follow the steps of Section~\ref{sec:gnn} and collected about $140k$ graphs.
We then randomly select $1,000$ graphs for the validation and $1,000$ graphs for the test and use the remaining graphs for training.

Meanwhile, we use the same set of commercially available molecules set $\mathcal{I}$ as~\cite{Chen2020,Kim2021}.
This set is build from \textit{eMolecules} (\url{http://downloads.emolecules.com/free/2019-11-01/}) that contains about $231M$ available molecules for our system.

\section{Baseline details}
$\bullet$ \textbf{Greedy DFS}~\citep{Hong2021}: This is a classic planning method that always prioritizes the node with max probability. We use tree-based Greedy DFS with $10$ as max search depth following previous works.

$\bullet$ \textbf{DFPN-E}~\citep{kishimoto2019depth}:  This is a  method based on Depth-First Proof-Number (DFPN) search.

$\bullet$ \textbf{MCTS-rollout}~\citep{Hong2021}: This is an MCTS method where nodes are molecules and edges are reactions.~\citep{Hong2021} is an MCTS method where nodes are molecules and edges are reactions.

$\bullet$ \textbf{EG-MCTS} and \textbf{EG-MCTS-0}~\citep{Hong2021}: They are MCTS methods with experienced guidance and achieved state-of-the-art results. The difference is that \textbf{EG-MCTS-0} does not use a value network.

$\bullet$ \textbf{Retro*} and \textbf{Retro*-0}~\citep{Chen2020}: They are tree-based A* search methods where Retro* using a value network but Retro*-0 not.

$\bullet$ \textbf{Retro*+} and \textbf{Retro*+-0}~\cite{Kim2021}: They improve the \textbf{Retro*} and \textbf{Retro*-0} method with self-training. 

\section{Reproducibility}

Our code is based on the Github repository of Retro* and Retro*+.
The policy GNN implementation is using the Pytorch Geometric~\citep{Fey/Lenssen/2019}.
The \uspto~data can be accessed from the original Retro* Github repository at \url{https://github.com/binghong-ml/retro_star}.
We use a NVIDIA TESLA P40 GPU for the model training and inference.
The search process also uses an Intel Xeon E5-2673 v3 CPU.

\end{document}